\documentclass[pmlr,twocolumn,square,numbers]{jmlr}
\jmlrproceedings{TerraBytes}{Proceedings of TerraBytes: Towards global datasets and models for Earth Observation Workshop at ICML}

\usepackage{longtable}
\usepackage{booktabs}
\usepackage{amsmath}
\usepackage{amssymb}
\usepackage{mathtools}
\usepackage[frozencache=true,cachedir=.]{minted}
\usepackage{caption}

\setcitestyle{square,numbers}

\title[Where are the Whales]{Where are the Whales: A Human-in-the-loop Detection Method for Identifying Whales in High-resolution Satellite Imagery}

\author{\Name{Caleb Robinson} \Email{caleb.robinson@microsoft.com}\\
\addr Microsoft AI for Good Research Lab, Redmond, Washington, USA
\AND
\Name{Kimberly T. Goetz}\\
\addr Marine Mammal Laboratory, Alaska Fisheries Science Center, National Marine Fisheries Service, NOAA, Seattle, Washington, USA
\AND
\Name{Christin B. Khan}\\
\addr Northeast Fisheries Science Center, National Marine Fisheries Service, NOAA, Woods Hole, Massachusetts, USA
\AND
\Name{Meredith Sackett}\\
\addr Azura Consulting, under contract to NOAA Fisheries, Northeast Fisheries Science Center, Woods Hole, MA USA
\AND
\Name{Kathleen Leonard}\\
\addr Protected Resources Division, Alaska Regional Office, National Marine Fisheries Service, NOAA, Anchorage, AK
\AND
\Name{Rahul Dodhia} \and \Name{Juan M. Lavista Ferres}\\
\addr Microsoft AI for Good Research Lab, Redmond, Washington, USA
}

\begin{document}

\maketitle

\begin{abstract}
Effective monitoring of whale populations is critical for conservation, but traditional survey methods are expensive and difficult to scale. While prior work has shown that whales can be identified in very high-resolution (VHR) satellite imagery, large-scale automated detection remains challenging due to a lack of annotated imagery, variability in image quality and environmental conditions, and the cost of building robust machine learning pipelines over massive remote sensing archives. We present a semi-automated approach for surfacing possible whale detections in VHR imagery using a statistical anomaly detection method that flags spatial outliers, i.e. ``interesting points''. We pair this detector with a web-based labeling interface designed to enable experts to quickly annotate the interesting points. We evaluate our system on three benchmark scenes with known whale annotations and achieve recalls of 90.3\% to 96.4\%, while reducing the area requiring expert inspection by up to 99.8\% --- from over 1,000 sq km to less than 2 sq km in some cases. Our method does not rely on labeled training data and offers a scalable first step toward future machine-assisted marine mammal monitoring from space. We have open sourced the entire pipeline at \url{https://github.com/microsoft/whales}.
\end{abstract}


\section{Introduction}
\label{sec:intro}

Understanding the spatial distribution of whales is essential for guiding conservation action, informing marine spatial planning, and upholding mandates such as the Marine Mammal Protection Act and the Endangered Species Act. The critically endangered North Atlantic right whale (\textit{Eubalaena glacialis}) has been experiencing a decline for over 10 years --- with an estimated 372 individuals remaining (95\% probability interval 360 to 383) as of 2024~\cite{hayes2024stock}. In the Pacific, the Cook Inlet beluga is also critically endangered, with an estimated population size of 381 individuals (95\% probability interval 317 to 473) as of 2022~\cite{goetz2023abundance}. Despite this, our ability to monitor these animals across broad spatial and temporal scales remains constrained by the cost and logistical complexity of traditional survey methods such as aerial, vessel-based, and acoustic monitoring.

Recent studies have shown that whales can be visually identified in very high-resolution (VHR) satellite imagery (e.g., 0.3 m/px or better)~\cite{fretwell2014whales,cubaynes2019whales,hodul2022individual}. These findings have led to exploratory work applying automated detection approaches, including deep learning models trained on satellite and aerial imagery datasets~\cite{borowicz2019aerial,guirado2019whale,boulent2023scaling}. Such models typically require large labeled datasets and do not generalize well to new environments, sensors, or ocean conditions. Recent efforts have consolidated all previous annotated satellite imagery into a single data archive~\cite{cubaynes2022whales}, however whale identification in novel satellite imagery is still largely driven by exhaustive expert annotation in desktop GIS software~\cite{khan2023biologist}. Manually annotating 100 km$^2$ of 31 cm/px imagery takes 3 hours and 20 minutes according to previous estimates~\cite{cubaynes2019whales}, limiting the feasibility of scaling these efforts as satellite imagery archives grow.

\begin{figure*}[t]
    \centering
    \includegraphics[width=1.0\linewidth]{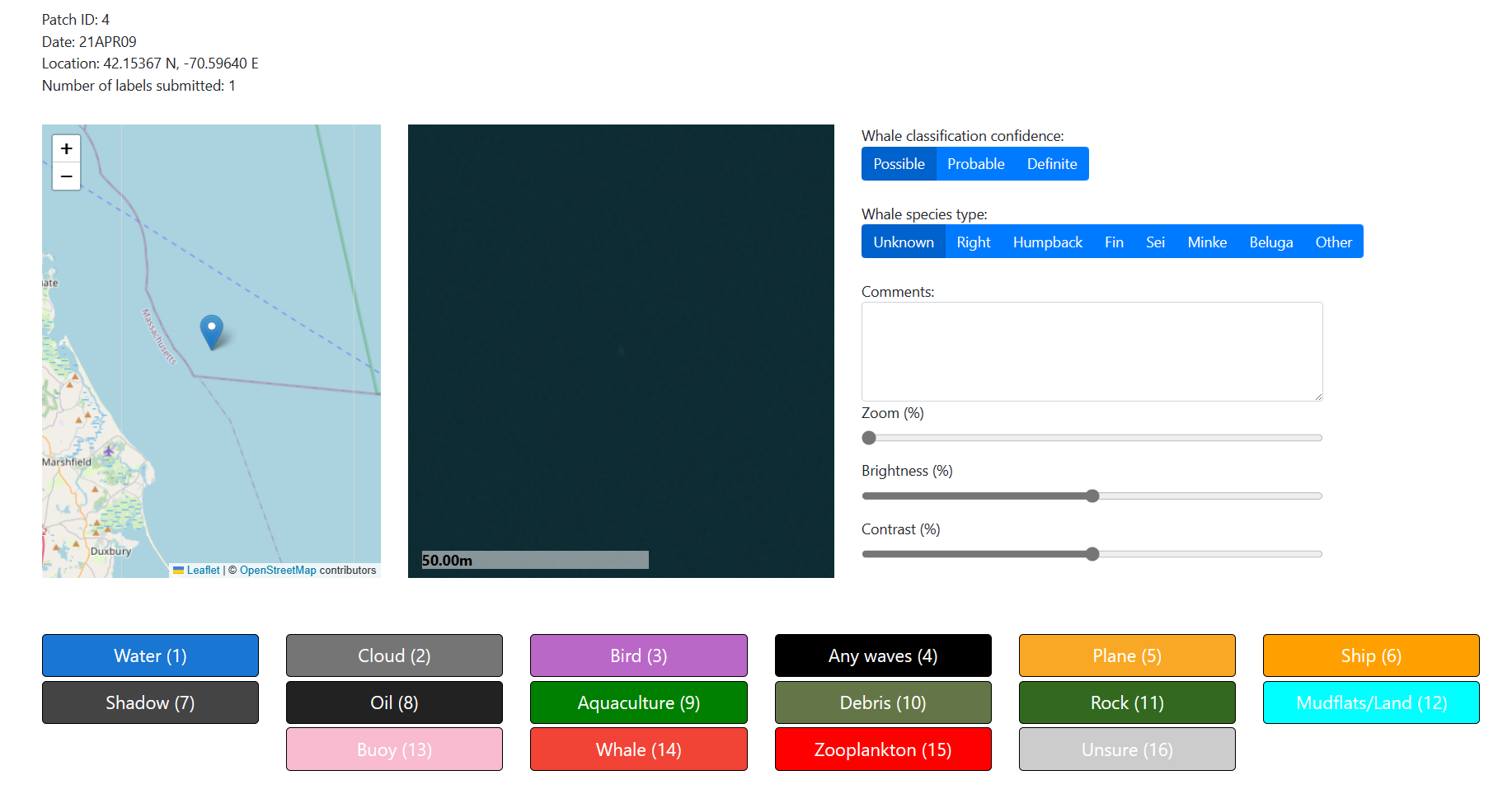}
    \caption{Web-based labeling application for quickly annotating ``interesting points''. For each candidate region of imagery flagged as statistically anomalous, labelers can inspect the corresponding imagery (the scale of each pixel is overlaid on the image), assign confidence levels (Possible, Probable, Definite), and classify object type (e.g., whale, ship, debris, zooplankton). Metadata such as location, date, and chips ID are displayed alongside adjustable image rendering controls (zoom, brightness, contrast). Once a class button is pressed the interface automatically loads a new image/interesting point to minimize the time per annotation.}
    \label{fig:interface}
\end{figure*}

In this work, we propose a simple alternative: using statistical anomaly detection to automatically identify outliers in VHR satellite scenes, and presenting these to experts through a browser-based labeling interface. Our method detects anomalous regions based on localized standardization of spectral intensities and aggregates these into a set of candidate detections -- ``interesting points''. These points can then be reviewed by human annotators, enabling more scalable annotation workflows and bootstrapping larger datasets for training deep learning models.

We evaluate our approach on three previously annotated Maxar VHR satellite scenes: from Cape Cod Bay in 2021~\cite{hodul2022individual}, Península Valdés in 2012~\cite{cubaynes2022whales}, and Península Valdés in 2014~\cite{cubaynes2022whales}, which collectively contain 174 manually annotated whales. Over imagery of calm water, our method achieves high recall while substantially reducing the search area that require manual inspection by an order of magnitude. While not a complete solution to whale detection, our system provides a useful first step in bridging the gap between manual annotation and future automated pipelines.

\section{Problem Formulation and Motivation}

We aim to develop a semi-automated framework for detecting whales in VHR optical satellite imagery of open water. Let $\mathcal{X}$ denote a large satellite image scene covering hundreds of square kilometers with sub-meter spatial resolution (e.g., $0.3\,\mathrm{m/px}$). The objective is to identify all observable marine mammals within $\mathcal{X}$ while minimizing the amount of expert labeling time required.

We assume that no prior dataset of annotated marine mammal instances is available for the specific region, ocean, or imaging conditions covered by $\mathcal{X}$. We do not assume access to a pre-trained vision model for detecting marine mammals in satellite imagery. Importantly, we assume that the imagery may be captured under non-ideal environmental conditions. The ocean surface may exhibit high texture variation due to wind-induced \textit{whitecaps}, and the scene may contain confounding artifacts such as vessels, buoys, aquaculture, floating debris, or even dense zooplankton swarms. Additionally, partial occlusion from clouds or haze may further complicate interpretation. Any practical method for solving this problem must therefore be robust to these sources of noise and variability, operating effectively under ``real-world'' constraints.

We frame the task of finding marine mammals in the ocean as one of anomaly detection. Our hypothesis is that marine mammals, while rare, will appear as local statistical outliers in an otherwise spatially homogeneous imagery\footnote{And further, imagery that doesn't fit this criteria -- e.g. images captured with a high frequency of white-capped waves -- will be identifiable by the relatively high of outliers and can be ignored.}. Using this assumption, we define a function $f_{\text{anomaly}}: \mathcal{X} \rightarrow \mathcal{P}$ that maps the input image $\mathcal{X}$ to a set of spatial locations $\mathcal{P} \subset \mathcal{X}$ corresponding to candidate points of interest. These points are selected based on deviations from local statistical norms.

The anomaly detection function $f_{\text{anomaly}}$ may be instantiated via classical statistical descriptors (e.g., local variance or z-scores within fixed-radius windows) or learned feature representations derived from deep models. The resulting candidate set $\mathcal{P}$ serves as a focus of attention for downstream expert review or targeted labeling, enabling a path for human-in-the-loop verification and training of object detectors in data-sparse settings.

\begin{figure}[t]
    \centering
    \includegraphics[width=0.9\linewidth]{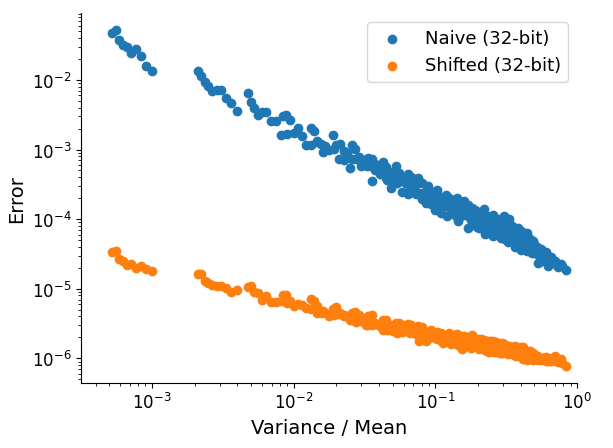}
    \caption{Comparison of numerical error in variance estimation under 32-bit precision for the naive and mean-shifted formulations of computing standard deviation in the \textbf{rolling window} approach. The $x$-axis shows the ratio of variance to mean intensity for synthetic image chips. The $y$-axis shows the absolute error in standard deviation relative to a double-precision ground truth. The naive formulation suffers from catastrophic cancellation when variance is small compared to the mean (which happens when using the rolling window approach over homogeneous ocean regions), while the shifted method achieves consistently lower error across all regimes.}
    \label{fig:catastrophic_cancellation}
\end{figure}

\section{Methods}

\subsection{Detection of Statistically Interesting Pixels}
Let \(\mathcal{X} \in \mathbb{R}^{C \times H \times W}\) denote a pansharpened and orthorectified VHR multispectral satellite image\footnote{Assuming trivial access to such imagery hides a huge amount of complexity, see \citep{goetz2025pixels} for an in-depth discussion of the topic. For the purposes of the discussion of anomaly detection methods we abstract this out, however `real-world' deployments of such methods require significant engineering.}, where \(C\) is the number of spectral channels and \(H \times W\) are the spatial dimensions. Our objective is to identify pixels in \(\mathcal{X}\) that exhibit statistically anomalous behavior relative to their local spatial context.

Our proposed method is simple -- we compute a per-pixel deviation score for each location in \(\mathcal{X}\), identifying extreme outliers pixels across the scene. The following sections describe two standardization methods for computing these deviation scores. High-deviation pixels are grouped into spatially contiguous regions and filtered based on geometric (i.e. size or outlier region) and statistical criteria. The resulting centroids of these regions, i.e. the ``interesting points'', form a candidate set \(\mathcal{P} \subset \mathcal{X}\) suitable for expert review or active labeling.

To focus the analysis and mitigate false positives, optional vector overlays (e.g., coastlines or land masks) can be applied to constrain the spatial domain. This is especially important in open-water applications, where land features can contain high-contrast structures that falsely trigger the anomaly detection mechanism. By masking out land and limiting analysis to known ocean extents, we increase the specificity of detection. Specifically, we use the Global Self-consistent, Hierarchical, High-resolution Geography Database (GSHHG)~\cite{wessel1996global} for land/water masking.

\subsection{Standardization Methods}

We compute a deviation tensor \(\mathcal{D} \in \mathbb{R}^{C \times H \times W}\) where each element \(D_{cij}\) captures the standardized anomaly of pixel \((i, j)\) in channel \(c\). We implement two methods for deriving \(\mathcal{D}\): \textbf{chunked standardization}, which uses coarse statistics, but is efficient to compute and \textbf{rolling window standardization} which uses local statistics, but is more computationally demanding.

\subsubsection{Chunked Standardization}

The first method partitions \(\mathcal{X}\) into non-overlapping square chunks of size \(s \times s\) (typically \(s=1024\) pixels). For each chunk, we compute the mean, \(\mu_c\), and standard deviation, \(\sigma_c\), for each channel. The deviation at each pixel is then computed as:
\[
D_{cij} = \frac{X_{cij} - \mu_c}{\sigma_c}
\]
This method is simple to parallelize and efficient (with time complexity \(\mathcal{O}(CHW)\)), but may overlook fine-grained deviations when anomalies are diluted across large homogeneous windows.

\subsubsection{Rolling Window Standardization}

The second method employs a sliding window approach that computes localized statistics centered at each pixel. For each location \((i,j)\), the local mean, \(\mu_{cij}\), and variance, \(\sigma_{cij}^2\), are estimated using a square window of size \(k \times k\) (e.g. \(k=31\) pixels):
\[
D_{cij} = \frac{X_{cij} - \mu_{cij}}{\sqrt{\sigma^2_{cij} + \varepsilon}}
\]
We implement this approach via channel-wise convolution operations using a uniform averaging kernel, see Listing \ref{lst:local_standardization} for a PyTorch implementation. This allows us to run the computation in parallel utilizing GPUs for processing.

\paragraph{Numerical stability considerations for the rolling window standardization.}

We estimate local variance using the identity \(\sigma^2_{cij} = \mathbb{E}[X^2] - (\mathbb{E}[X])^2\), as both terms can be implemented as convolutions. However, in regions where the variance is small (e.g., calm water), this formulation suffers from catastrophic cancellation due to the limited precision of 32-bit floating point arithmetic. To address this, we subtract a global per-channel mean, \(\bar{X}_c\), from \(\mathcal{X}\) before convolution:
\[
X'_{cij} = X_{cij} - \bar{X}_c
\]
This is a standard method for stabilizing the computation of \(\sigma^2\) by reducing the absolute magnitude of the squared and squared-mean terms. Figure \ref{fig:catastrophic_cancellation} shows the effect that this \textit{shifted} method has on the error of the computation as a function of the similarity of the variance and mean. Finally, we add a small constant, \(\varepsilon = 10^{-8}\), to the denominator to ensure positive values and avoid undefined behavior under the square root.

\subsection{Anomaly Aggregation and Feature Extraction}

We aggregate the channel-wise deviations, \(\mathcal{D}\), across a large input scene over the channel dimension in a scalar anomaly map, \(A \in \mathbb{R}^{H \times W}\), as follows:
\[
A_{ij} = \sum_{c=1}^{C} |D_{cij}|
\]
We then threshold \(A\) either at a fixed value or using a high quantile (e.g., the 99.99th percentile) to produce a binary mask of anomalous pixels. We then extract spatially contiguous connected components (using 8-connected neighborhood logic) and compute their average anomaly intensity.

We filter out regions whose area falls below a minimum threshold (e.g., less than 1.5 square meters). We then represent surviving features as points using their geometric centroids and record their area and average aggregate deviation value.

\begin{figure}[th]
    \centering
    \includegraphics[width=0.9\linewidth]{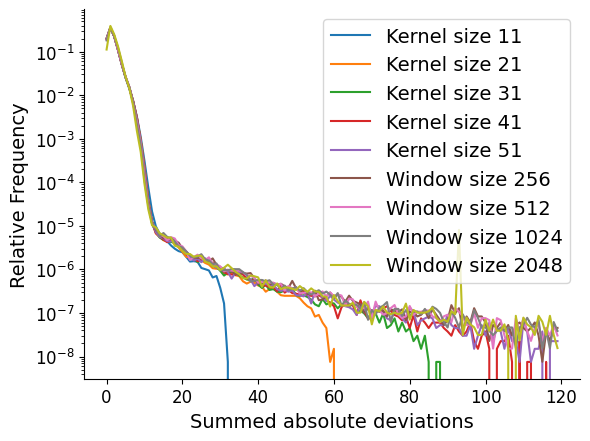}
    \includegraphics[width=0.9\linewidth]{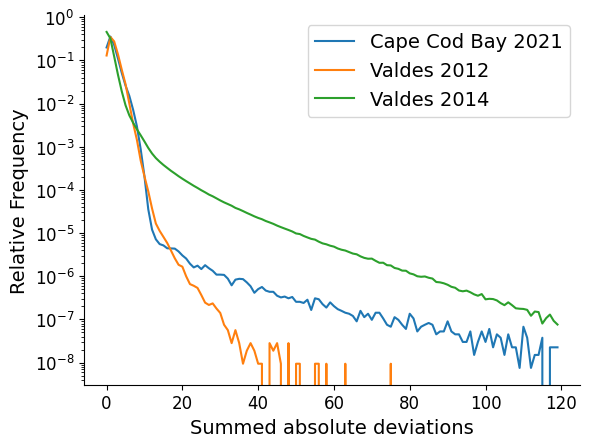}
    \caption{(\textbf{Top}) The distribution of anomaly values (summed absolute deviations) on the Cape Cod Bay scene~\cite{hodul2022individual} using the two standardization methods with varying kernel ($k$) or window ($s$) sizes. We find that both methods result in similar distributions with a window or kernel size greater than 50. (\textbf{Bottom}) Comparison of anomaly value distributions (using rolling window standardization with a $k=51$ kernel) across the three evaluation scenes. The Valdés 2014 scene has a high density of high-contrast features (mostly white-capped waves) which shows up as a longer tail in the anomaly distribution.}
    \label{fig:deviations}
\end{figure}

\begin{table*}[th]
\centering
\resizebox{\textwidth}{!}{%
\begin{tabular}{lcccc}
\toprule
\textbf{Scene} & \textbf{Annotated Whales} & \textbf{True Positives} & \textbf{False Positives} & \textbf{Recall} \\
\midrule
Cape Cod Bay 2021 & 31 & 28 & 280 & 90.3\% \\
Península Valdés 2012 & 84 & 81 & 276 & 96.4\% \\
Península Valdés 2014 & 59 & 54 & 3622 & 91.5\% \\
\bottomrule
\end{tabular}
}
\caption{Detection performance across three benchmark scenes.}
\label{tab:results}
\end{table*}

\section{Experiments and Results}

Our proposed methods have several free parameters that influence the quality and number of ``interesting points'' that a labeler must examine: the size of the window over which statistics are computed ---  $s$ in the \textbf{chunked standardization} approach and $k$ in the \textbf{rolling window standardization} approach --- the anomaly threshold used for binarizing $A$ (determining which pixels are anomalous), and the area threshold (determining how large an anomalous region must be to be considered ``interesting''). A desirable configuration is one in which all of the whales in a scene are marked as ``interesting'' (i.e. high recall) and few non-whale points are marked as ``interesting'' (i.e. high precision).

We evaluate the effectiveness of our methods by applying them to three satellite scenes previously studied in the literature, each containing manually annotated whale locations:
\begin{description}
    \item[Cape Cod Bay 2021~\citep{hodul2022individual}] Two WorldView-3 scenes (Catalog IDs 1040010067D36B00 and 10400100674B2100) captured on April 24, 2021 covering $\sim$200 km$^2$ of Cape Cod Bay\footnote{We approximate the study area used in \citep{hodul2022individual} by buffering the extent of the annotated points by 500m.} at a 0.3 m/px spatial resolution. It contains 31 annotated North Atlantic right whales, 25 of which are considered definite, and 6 of which are ambiguous.
    \item[Península Valdés 2012~\cite{cubaynes2022whales}] A WorldView-2 scene (Catalog ID 103001001C8C0300) captured on September 19, 2012, covering 120 km$^2$ of the Valdés Peninsula in Argentina with a spatial resolution of 0.56 m/px. It includes 84 Southern right whales (\textit{Eubalaena australis}) labels categorized by confidence: 15 definite, 32 probable, and 37 possible sightings.
    \item[Península Valdés 2014~\cite{cubaynes2022whales}] A WorldView-3 scene (Catalog ID 10400100032A3700) captured on October 16, 2014 covering 560 km$^2$, also from the Valdés Peninsula, with a spatial resolution of 0.37 m/px. It includes 59 Southern right whale labels: 23 definite, 12 probable, and 24 possible.
\end{description}

\paragraph{Window size} We run the chunked and rolling window standardization methods with different window sizes for the Cape Cod Bay 2021 scene and show the distribution of aggregated anomaly scores in the top panel of Figure \ref{fig:deviations}. We find similar distributions of anomaly values among window sizes of 256, 512, 1024, and 2048 for the chunked standardization method and kernel sizes of 41 and 51 for the rolling window method. Kernel sizes of less than 41 see a drop-off in tail values as there is less context with which to determine whether a given value is an anomaly or not. Practically, on an V100 GPU, there is little difference in execution time of the rolling window standardization computation with larger window sizes.

\paragraph{Threshold values} We run the rolling window standardization method with $k=51$ on each of the three evaluation scenes and plot the distribution of anomaly values in the bottom panel of Figure \ref{fig:deviations}. We find that the distributions from the Cape Cod Bay 2021 and Valdés 2012 scenes --- both of which have very still water --- have shorter tails, while the Valdés 2014 scene --- with a large number of white-capped waves --- has a long tailed distribution with 99.99th percentile values of 11.37, 11.90 and 55.25, respectively (computed over the RGB channels only). We always use a conservative value of 1.5 square for area thresholding based on conversations with experts and observing that positive identifications may only highlight parts of a whale. In general, we find that it is possible to run the interesting point methods with a range of anomaly thresholds and choose the result that returns a \textit{reasonable} number of points per square kilometer after area filtering (using $<2$ points/sq km as a rough cutoff).

\paragraph{Results} We evaluate our methods using the 99.99th percentile anomaly threshold for Cape Cod Bay 2021 and Valdés 2012 and the 99.9th percentile anomaly threshold for Valdés 2014 (trading off a larger number of false positives for higher true positives). We define an ``interesting point'' as a true positive if it falls within a 100 meter radius of an annotated whale location and a false positive otherwise. Results are summarized in Table~\ref{tab:results}. We find that most waves are identified as anomalous -- `false positives'. In scenes with still water, we find recall values greater than 90\% with relatively few false positives.

\begin{figure*}[thbp]
    \centering
    \includegraphics[width=1.0\linewidth]{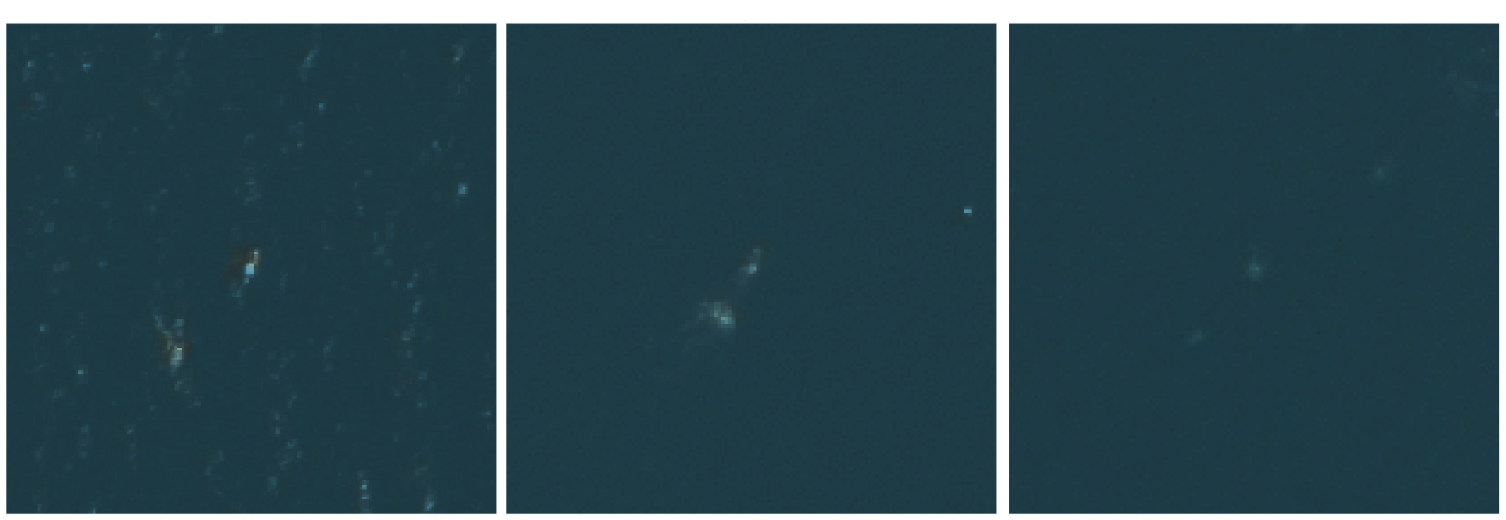}
    \caption{Example whale detections from the May 22, 2020 scene (\textbf{left}) and two detections from the larger Cape Cod Bay 2021 scene that were not considered in the Hodul et al. 2022 study (\textbf{middle, right}). The May 2020 scene (1,056 sq km) produced 220 interesting points, each of which were labeled by 3 experts in a total of 31 minutes. The Cape Cod Bay scene (1,083 sq km) produced 555 interesting points, each was also labeled by 3 experts in a total of 139 minutes.}
    \label{fig:whales}
\end{figure*}

\begin{figure}
    \centering
    \includegraphics[width=1.0\linewidth]{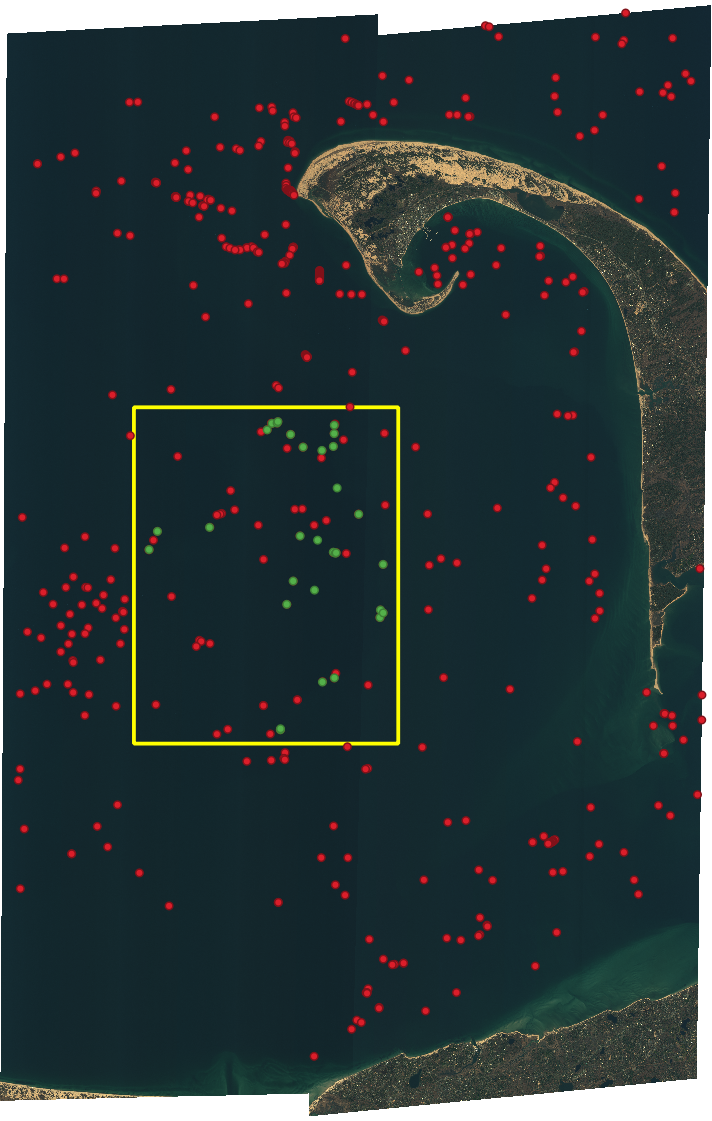}
    \caption{A map of 555 ``interesting points'' in \textcolor{red}{red} found over the two WorldView-3 scenes from used in the \citep{hodul2022individual} study. The whale points from the Hodul et al. study are shown in \textcolor{green}{green} and the study area (buffered from the extent of the labeled points) is shown in \textcolor{yellow}{yellow}. Each of the ``interesting points'' was annotated by three expert labelers which took a total of 139 minutes. Manual annotation of the entire scene would take approximately $\sim 35$ hours. The interesting points covered 28 out of the 31 whales found in the Hodul et al. study (which were all correctly identified by the labelers). Further, 5 additional whale points were found outside of the previously examined study area.}
    \label{fig:hodul}
\end{figure}

\section{Labeling interface}

To facilitate rapid annotation of ``interesting point'' detections in large amounts of satellite imagery, we developed a browser-based labeling tool designed for expert reviewers -- shown in Figure \ref{fig:interface}.

Each session presents the user with a sequence of $100 \mathrm{m} \times 100\mathrm{m}$ image \textit{chips} centered on geographic centroids of the interesting regions outputted from the anomaly detection pipeline. These chips are accompanied by contextual metadata, including the imagery acquisition date and geographic coordinates, which are also visualized on an interactive map. Users can select one of sixteen predefined semantic classes (e.g., \texttt{whale}, \texttt{ship}, \texttt{debris}, \texttt{oil}), and for whale detections also assign species, confidence (\textit{possible}, \textit{probable}, or \textit{definite}), and add free-form comments. After a class is selected, the interface will immediately load the next chip, facilitating rapid review.

The tool provides controls for adjusting image zoom, brightness, and contrast in real time. These settings allow users to optimize visual clarity when interpreting difficult scenes, such as those with haze, glint, or low contrast.

The labeling system itself is also deliberately simple and is implemented as a single multi-threaded Python HTTP server application that operates in a stateless mode with a single page web frontend. No login credentials are required, and labeler identity is tracked solely through self-supplied ids to avoid duplicate sampling. Each chips is assigned to multiple labelers and is automatically removed from circulation once it has been annotated by three distinct labelers. This redundancy provides a simple mechanism for quality control through label consensus, while allowing for validation in downstream workflows. All labels and associated metadata are saved in a CSV file by the server application.

The anomaly detection pipeline, labeling interface and associated backend are open-sourced with demo data and setup instructions at \url{https://github.com/microsoft/whales}.

\begin{listing*}[t!]
\begin{minted}[fontsize=\small, breaklines]{python}
class LocalContextStandardization(Module):
    def __init__(self, in_channels: int = 3, kernel_size: int = 9, shift_val=None):
        super().__init__()

        self.shift_val = shift_val

        weights = torch.nn.Parameter(
            torch.zeros(
                in_channels, in_channels, kernel_size, kernel_size, dtype=torch.float32
            ),
            requires_grad=False,
        )
        for i in range(in_channels):
            weights[i, i] = (
                torch.ones(kernel_size, kernel_size, dtype=torch.float32)
                / kernel_size**2.0
            )

        self.conv = Conv2d(
            in_channels,
            in_channels,
            kernel_size=kernel_size,
            padding="same",
            padding_mode="replicate",
            bias=False,
        )
        self.conv.weight = weights

    def forward(self, x: Tensor) -> Tensor:
        if self.shift_val is not None:
            x = x - self.shift_val
        else:
            x = x - x.mean(dim=(0, 2, 3), keepdim=True)
        mu = self.conv(x)
        squares = self.conv(x**2.0)
        variance = squares - mu**2.0
        return (x - mu) / (torch.sqrt(variance) + 1e-8)
\end{minted}
\caption{Implementation of the \textbf{rolling window standardization} approach, which applies a channel-wise local standardization to imagery using convolution-based estimates of mean and variance.}
\label{lst:local_standardization}
\end{listing*}

\section{Case Study and Discussion}

To assess the practical deployment of our system in new, unlabeled imagery, we applied the anomaly detector and labeling interface to a previously unstudied scene captured on May 22, 2020, over Cape Cod Bay (WorldView-3 catalog ID 10400100585EFA00). The scene covers 1,056 sq km at a 0.3 m/px resolution and generally has calm water throughout the scene.

We used the rolling window standardization approach with deviation threshold of the 99.99th percentile and an area threshold of 1.5 sq meters which found 220 ``interesting points'', corresponding to approximately 0.21\% of the total pixels\footnote{Calculated by using a 50m buffer around each interesting point which corresponds to the area shown to the labeler in the user interface.}. Each of these were subsequently annotated by three expert labelers using our the user interface which took 31 minutes of total human effort (as measured by the total time between the interface serving an image chip to be labeled and the subsequent response with class annotation over the three labelers). Using the approximation from \citep{cubaynes2019whales} --- that manually annotating 100 sq km of imagery takes approximately 3 hours and 20 minutes --- annotating this scene would have taken a single labeler $\sim 35$ hours.

Most detections were attributed to \texttt{whitecap} (n=420) and \texttt{unsure} (n=234), reflecting the presence of breaking waves near the shore. Only two points were annotated as a  \texttt{whale}, of which, all three labelers agreed on one point as a whale (see Figure \ref{fig:whales}). We observe that the \texttt{whale} annotations took substantially longer to process --- ranging from 15 to 106 seconds each --- due to increased scrutiny and zooming behavior by the annotators. In contrast, non-whale categories required only 2.8 seconds per chips on average (with a standard deviation of 8.5 seconds).

Further, we applied the same procedure to the two scenes used in the Hodul et al. 2022 study~\cite{hodul2022individual} (see the Cape Cod Bay 2021 description). This study focused on a $\sim$200 sq km area taken from two larger WorldView-3 strips, however the entire area from the two strips covers 1,083 sq km. Figure \ref{fig:hodul} shows this larger extent, with the approximate study area shown as a yellow box. We find a total of 555 interesting points in the larger scene, which took expert labelers 139 minutes in total to annotate. The average time per non-whale annotation was 5.1 seconds while the average time per \texttt{whale} annotation was 11.23 seconds. The 555 interesting points covered 28 out of the 31 interesting points found in the Hodul et al. study, and the annotators correctly annotated each as a whale. Additionally, the annotators flagged 5 other points outside of the original study area as whales, two of which are shown in Figure \ref{fig:whales}. In both cases, the cost of pre-processing the scenes and applying our methods is negligible compared to the cost of having marine biologists annotate them.

Finally, we observe -- and emphasize as a limitation -- that applying our methods to scenes that contain white-capped waves from high winds results in a large number of false positives. The simple methods we propose here are not able to distinguish between different classes of anomalous groupings of pixels (i.e. between whitecaps, whales, buoys, or otherwise), and further work and labeled datasets are needed to find whales in these challenging conditions. Nevertheless, satellite imagery archives contain unstudied imagery of calm seas that can be mined for whale detections and used to bootstrap larger modeling efforts.

\section*{Impact Statement}
This work contributes tools and methodology for identifying whales in satellite imagery to support marine mammal monitoring and conservation at scale. By reducing the need for exhaustive manual annotation, our semi-automated pipeline enables domain experts to focus their attention on anomalous ``interesting points'', accelerating detection of whales in vast ocean areas.

We emphasize that these methods are not fully automated and require expert interpretation to ensure reliable use. The approach is sensitive to imaging conditions, scene complexity, and spectral variability, and may produce false positives in challenging environments. As such, it is intended as a decision-support system to augment --- not replace --- expert-driven analysis.

All system components are open-sourced to promote transparency, reproducibility, and community adoption. This work marks a key step toward scalable, expert-in-the-loop remote sensing for conservation. By enabling rapid review of vast ocean imagery for spectrally distinct features, it allows experts to label whales and other surface objects, creating high-quality training data. This data can power future model development, accelerating geospatial insights into whale presence and enhancing both conservation efforts and maritime domain awareness.

\bibliography{citations}

\end{document}